\documentclass{article}
\usepackage{iclr2026_conference,times}

\usepackage{amsmath,amsfonts,bm}

\def\eqref#1{equation~\ref{#1}}

\def\1{\bm{1}}

\DeclareMathAlphabet{\mathsfit}{\encodingdefault}{\sfdefault}{m}{sl}
\SetMathAlphabet{\mathsfit}{bold}{\encodingdefault}{\sfdefault}{bx}{n}

\usepackage{graphicx}
\usepackage{booktabs}
\usepackage{xcolor}
\usepackage{enumitem}
\usepackage{hyperref}
\usepackage{url}

\usepackage[accsupp]{axessibility}

\newcommand{\eg}{e.g.}
\newcommand{\ie}{i.e.}

\newcommand{\etal}{et al.}

\title{Last-Meter Precision Navigation for UAVs: A Diffusion-Refined\\ Aerial Visual Servoing Approach}

\newcommand{\authorblock}[1]{%
  \makebox[0.5\textwidth][l]{%
    \normalfont
    \begin{tabular}[t]{@{}l@{}}
      #1
    \end{tabular}%
  }%
}
\renewcommand{\headrulewidth}{0pt}
\author{%
\authorblock{
\textbf{Yaxuan Li}\,\textsuperscript{*} \\
FST and ICI, University of Macau \\
Macau SAR, China \\
\texttt{yaxuanli.cn@gmail.com}
}
\And
\authorblock{
\textbf{Jiarui Zeng}\,\textsuperscript{*} \\
FST and ICI, University of Macau \\
Macau SAR, China \\
\texttt{zeng.jiarui@connect.um.edu.mo}
}
\AND
\authorblock{
\textbf{Shaofei Huang} \\
FST and ICI, University of Macau \\
Macau SAR, China \\
\texttt{nowherespyfly@gmail.com}
}
\And
\authorblock{
\textbf{Zhedong Zheng}\,\textsuperscript{\ddag} \\
FST and ICI, University of Macau \\
Macau SAR, China \\
\texttt{zhedongzheng@um.edu.mo}
}
}

\iclrfinalcopy

\begin{document}

\maketitle

\renewcommand{\thefootnote}{\fnsymbol{footnote}}
\footnotetext[1]{Equal contribution.}
\footnotetext[3]{Corresponding author.}

\renewcommand{\thefootnote}{\arabic{footnote}}
\setcounter{footnote}{0}

\begin{abstract}
In this work, we study the last-meter precision navigation for UAVs, \eg, autonomously reaching a target within the final 10 meters using monocular vision. This task is challenging due to scale ambiguity, rotation discontinuities, and the need for fine-grained spatial reasoning. Existing methods often fail under large viewpoint changes or lack generalization to unseen environments. To this end, we propose \textbf{DreamNav}, a coarse-to-fine diffusion-refined aerial visual servoing framework. 
In the first coarse-estimation stage, a robust regression policy employs a trigonometric parameterization to predict rotation by jointly modeling sine and cosine components, effectively mitigating optimization instabilities caused by angular periodicity. 
Given this coarse estimate, the second diffusion-refined stage utilizes a pre-trained world model to simulate future visual observations for candidate actions, selecting the trajectory that minimizes visual discrepancy with the target through a process of visual imagination. To support rigorous evaluation, we contribute \textbf{PairUAV}, a large-scale benchmark comprising 4.8 million image pairs across 72 scenes, curated from the University-1652 dataset.
Extensive experiments show DreamNav outperforms strong visual servoing and foundation model baselines in accuracy and generalization, with zero-shot transfer to unseen scenes.\footnote{
Code is available at: \url{https://github.com/YaxuanLi-cn/PairUAV.git}.\\
Dataset is available at: \url{https://huggingface.co/datasets/YaxuanLi/pairUAV/tree/main}.
}

\end{abstract}

\section{Introduction}
\label{sec_intro}

\begin{figure}[!t]
    \centering
    \begin{minipage}[t]{0.7\linewidth}
        \vspace{0pt}
        \centering
        \includegraphics[width=\linewidth]{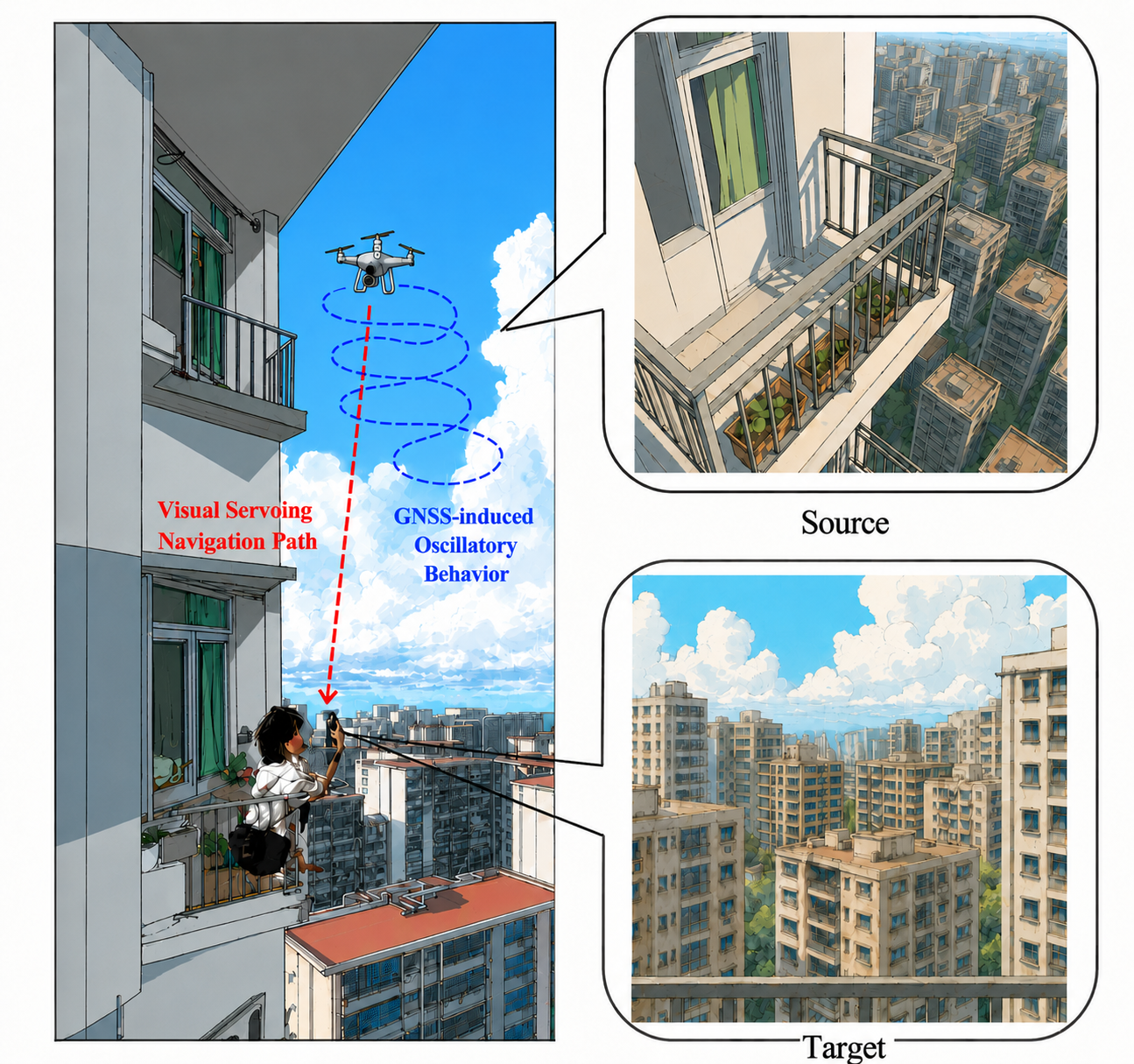}
    \end{minipage}
    \hfill
    \begin{minipage}[t]{0.25\linewidth}
        \vspace{0pt}
        \caption{
        \textbf{Last-Meter Precision Navigation.}
        (\emph{Left}) Many drone applications, such as package delivery, necessitate high-precision localization during \textcolor{red}{the final approach}, a phase where \textcolor{blue}{GNSS signals} are often unreliable or unavailable, particularly for resolving vertical position.
        (\emph{Right}) To overcome this challenge, we frame the task as a visual servoing problem, where the drone estimates the relative pose between its current view and a target image to guide its trajectory to the final destination based on visual refinement.
        }
        \label{fig:motivation}
    \end{minipage}
\end{figure}

The ``last-meter'' precision navigation scenario is critical for a wide range of real-world applications, including disaster search and rescue~\cite{PATKI2022102558, 9155522}, swarm organization~\cite{kramaric2025comprehensive, bu2024advancement} and autonomous delivery~\cite{10.1108/IJCHM-07-2018-0558, brunner2019urban}. Despite the maturity of global navigation satellite systems (GNSS), their accuracy is typically limited to several meters and lacks the ability to resolve fine vertical distinctions or discriminate between closely spaced structures (\eg, windows or balconies)\footnote{See official GPS performance standards\cite{gpsps2020}.}. As a result, the primary challenge in aerial navigation has shifted from coarse global localization to fine-grained local positioning and precise terminal pose alignment (see Figure~\ref{fig:motivation}).  

Despite recent advances in aerial navigation~\cite{7989381, deitke2020robothor}, two critical challenges remain unresolved in achieving robust last-meter navigation for UAVs. \textbf{(1)} There is a lack of large-scale datasets tailored to close-range aerial navigation, particularly those capturing fine-grained spatial variations at different distances and viewpoints. Existing benchmarks often focus on long-range localization or indoor flight, failing to model the visual dynamics encountered during terminal descent and alignment. 
\textbf{(2)} Accurately estimating subtle differences in relative pose remains highly challenging, especially under scale ambiguity (\eg, meter-level displacements) and rotational discontinuities (\eg, small angular deviations). This challenge is further exacerbated by varying lighting and viewpoint conditions, which demand robust, fine-grained spatial reasoning. This, in turn, requires models to extract discriminative visual features and perform precise metric regression.
Current approaches~\cite{deuser2023sample4geo} struggle under large viewpoint changes and exhibit poor generalization to unseen environments (see Section~3), primarily due to their reliance on direct regression without explicit modeling of visual dynamics.

(1) For the first challenge, we introduce PairUAV, the first large-scale open-world benchmark for image-driven UAV navigation. PairUAV spans 72 geographic locations and 1,652 building-centric targets, comprising over 4.8 million navigation instances that capture the semantic diversity and visual complexity of real environments. In each trial, the UAV is initialized several hundred meters away from the target, obtains egocentric RGB observations from a downward-facing sensor, and the dataset provides ground-truth rotation and displacement labels to reach the goal.

(2) For the second challenge, we argue that visual servoing offers a promising solution by enabling dynamic, feedback-driven refinement of positional accuracy. While visual servoing has been extensively studied in controlled environments, such as industrial automation and factory assembly systems~\cite{7989381}, and in robotic manipulation~\cite{bar2025navigation, ren2025prior}, its application to unmanned aerial vehicles (UAVs) in complex outdoor environments remains significantly underexplored.  
In particular, we propose DreamNav, a coarse-to-fine, diffusion-refined visual servoing framework. In Stage I, instead of regressing the rotation angle directly, we introduce a trigonometric loss that jointly constrains cosine and sine components to accurately regress the rotation angle, thereby avoiding angle-wrapping issues. Building on this loss, a ViT-based backbone is fine-tuned to predict a rotation angle and the corresponding forward distance. In Stage II, we leverage a large-scale world model to imagine the visual outcome of each candidate action and select the one whose predicted view best matches the target image. Unlike direct prediction, which easily overfits to narrow, brittle cues (\eg, the edge of a square building) that may vanish or be misdetected in new scenes, imagination enforces image‑wide consistency, requiring the policy to account for scene geometry, semantics, and viewpoint changes across the entire frame. This holistic constraint penalizes single‑cue reliance, provides a built‑in test‑time self‑verification signal, and thus improves robustness under distribution shift, yielding stronger generalization to unseen scenes. 

To enable systematic evaluation, we establish a standardized protocol on our proposed PairUAV benchmark. We assess all methods on unseen testsets, featuring novel environments to evaluate out-of-distribution generalization. Performance is quantified using three aggregate metrics averaged over all episodes: mean angular error, mean distance error, and success rate. We benchmark our method, DreamNav, against strong baselines from two distinct families: vision servoing models (AI2THOR), vision foundation models (Dinov3-ViT7B, Sample4Geo). In brief, our primary contributions are summarized as follows:
\begin{itemize}[leftmargin=2em,itemsep=0.25em]
    \item \textbf{A Large-Scale Benchmark for Last-Meter Navigation.} We identify the lack of standardized evaluation for fine-grained, close-range UAV navigation, particularly in the critical final 10 meters. To fill this gap, we introduce \textbf{PairUAV}, a comprehensive benchmark featuring 4,817,232 image pairs across 72 diverse scenes, curated from real-world environments. It enables rigorous evaluation of visual servoing methods under scale ambiguity, viewpoint variation, and fine spatial reasoning.

    \item \textbf{A Coarse-to-Fine Diffusion-Refined Framework.} We propose \textbf{DreamNav}, a novel visual servoing framework that combines robust coarse estimation with diffusion-based refinement. In the first stage, a trigonometrically parameterized policy ensures stable rotation prediction. In the second stage, DreamNav leverages a pre-trained world model and a diffusion-inspired imagination process to simulate future observations and select optimal actions, effectively refining pose estimates through visual prediction.

    \item \textbf{Superior Accuracy and Generalization in Realistic Settings.} Extensive experiments show that DreamNav achieves state-of-the-art performance on the PairUAV benchmark, with mean absolute error of \textbf{38.78} in heading prediction and \textbf{29.16} in range prediction. It significantly outperforms both classical visual servoing baselines and powerful foundation models, showing strong zero-shot generalization to real scenes and robustness to large viewpoint changes.
\end{itemize}

\section{Related Work}
\label{related_work}

\noindent \textbf{Drone Navigation.} Drone navigation has received increasing attention in recent years, propelled by the proliferation of Unmanned Aerial Vehicles (UAVs) as a powerful tool for gathering rich and diverse multimedia content. Their unique vantage point enables applications in aerial photography~\cite{zhao2023ms,shao2022efficient}, cinematography~\cite{pitas2022autonomous,dang2022path}, and geo-localization~\cite{liu2019lending,shi2019spatial,shi2020optimal,shi2020looking,shi2022beyond,shi2022accurate,shi2022cvlnet,shi2023boosting,song2023learning}, among others. Early navigation methods often assume reliable GNSS and strong localization, which is extremely challenging in signal-denied settings. To address this, recent work guides UAVs to their targets by leveraging richer onboard inputs, a direction further empowered by advancements in computer vision and machine learning~\cite{wang2022multiple,wang2022learning,zheng2022adaptive}. For example, some approaches cast movement commands into semantic, instruction-following formulations~\cite{xiao2025uav, chu2024towards}, while others specify the destination via detailed goal descriptors~\cite{liu2023aerialvln}. In this work, we guide a UAV for last-meter precision navigation using a single target-view image in lieu of GNSS. Unlike richer multimodal or map-based inputs, this single-image cue is lightweight and easy to acquire yet provides less explicit information, making the problem more challenging and practically relevant.

\noindent \textbf{Vision Servoing Models.} Visual servoing uses visual feedback in a closed loop to command robot motion. Two canonical formulations are Position-Based Visual Servoing (PBVS) and Image-Based Visual Servoing (IBVS). Recently, deep neural networks have shown strong performance in this area, benefiting from their capability in automated image and video analysis~\cite{liu2023robust,li2021uav,9854892,zheng2019unsupervised,wang2023context,ju2024video2bev}. For instance, Zhu~\etal~\cite{7989381} first apply deep neural networks to IBVS, and subsequent work~\cite{ren2025prior, bar2025navigation} expands the data and substantially improved performance. Our approach follows this line by injecting target image cues into a neural network to close the loop, but with two key differences. First, prior studies focus on ground navigation, whereas we address aerial navigation, a domain of growing importance for capturing data from previously inaccessible or hard-to-reach locations~\cite{zhang2019eye,yuanqiang2020guided,liu2023robust}. To the best of our knowledge, this is the first visual-servoing system studied for aerial scenarios. Second, earlier approaches treat decisions as classification on a finite action set, while we instead learn a regressor that outputs continuous, parameterized control.

\setlength{\tabcolsep}{5pt}

\begin{table}[!t]
\centering
\caption{Comparison with the existing ObjectNav and Aerial Navigation benchmarks.}
\label{tab:benchmark_compare}
\small
\setlength{\tabcolsep}{3pt} 
\begin{tabular}{@{}l| l | c | c c @{}}
\toprule
Benchmark & Agent & External Inputs & \#DoF & \#Instances \\
\midrule
AI2-THOR \cite{7989381} & Robot & Target Image & - & 2,176 \\
RoboTHOR \cite{deitke2020robothor} & Robot & Object Category & 6 actions & 731 \\
UAV-ON \cite{xiao2025uav} & Drone & Target Description & 4 & 11,000 \\
AerialVLN \cite{liu2023aerialvln} & Drone & Movement Instr. & 4 & 25k \\
GeoText \cite{chu2024towards} & Drone & Target Description & - & 276k \\
\midrule
PairUAV (ours) & Drone & Target Image & 2 & 4,817,232 \\
\bottomrule
\end{tabular}
\end{table}


\section{PairUAV Dataset}
\label{PairUAV}

As discussed in Sec.~\ref{sec_intro}, the last‑meter precision navigation task is of substantial practical relevance yet remains underexplored. Given that, we begin by formalizing the problem, specifying the problem setup, input–output spaces, and evaluation metrics. We then present our PairUAV, a large‑scale dataset curated for this task, describe its collection and annotation pipeline, and compare it against existing datasets in terms of scene coverage, labeling granularity, and difficulty distribution.

\subsection{Task Definition}
We formulate the problem of image-based terminal aerial navigation in open-world environments. At the start of each episode, a UAV is initialized at a starting viewpoint and tasked with navigating to a target viewpoint defined by visual observations. The UAV is equipped with an onboard monocular RGB camera capturing images at a resolution of $512\times512$. An episode is considered successful if the UAV reaches the vicinity of the target location within a distance threshold of $d_{\text{succ}}=10$ meters, following standard protocols~\citep{xiao2025uav}.

To enhance flight stability and mitigate the perceptual ambiguity associated with monocular depth, we adopt a simplified 2-DoF navigation interface. Formally, we fix the camera pitch at $\psi=45^\circ$ and maintain a constant horizontal field of view, eliminating the need for the agent to reason about pitch dynamics or zoom variations. Under these constraints, the relative displacement between viewpoints is fully characterized by (i) a change in heading (yaw), and (ii) a displacement along the camera's viewing direction. Consequently, we parameterize the navigation action space as two continuous scalars:
\begin{equation}
a = [\Delta \theta, \Delta r],
\end{equation}
where $\Delta \theta$ denotes the heading adjustment and $\Delta r$ represents the change in range to the target. This formulation effectively captures the essential terminal behavior of ``turn-to-align and move-to-approach'' while abstracting away altitude, pitch, and other complex motion components.

\subsection{Dataset construction}
We build upon the University-1652 dataset~\cite{zheng2020university}, adopting its established protocol and 3D environments, which span 1,652 buildings across 72 universities. The original protocol samples 54 proximal coordinates for each building, enforcing a minimum vertical offset and a maximum inter-sample radius to simulate UAV operation in GNSS-denied settings (\ie, short lateral baselines with substantial altitude variance). Monocular UAV views are rendered from Google Earth models at each coordinate. Our primary contribution deviates from the original dataset's cross-view retrieval task. We repurpose these assets for precision navigation. Specifically, for each building, we generate all ordered pairs from the 54 sampled coordinates. For each pair, we compute and store the precise 2-DoF relative pose (range translation and rotation) transforming the first view to the second. This methodology yields a new, large-scale dataset of 4,817,232 ordered navigation instances, providing explicit supervision for relative camera pose estimation.

\subsection{Comparison with other datasets}


In Table~\ref{tab:benchmark_compare}, we compare \textsc{PairUAV} with established ObjectNav and aerial navigation benchmarks.
Compared to AI2-THOR and RoboTHOR, which evaluate ground robots conditioned on category labels or category sequences, \textsc{PairUAV} targets image-goal aerial navigation.
In contrast to category- or text-specified targets (e.g., category labels, movement instructions, or descriptions), \textsc{PairUAV} specifies the goal by an RGB image.
This design aligns the supervision with the control objective of reaching the target viewpoint.
On the action side, \textsc{PairUAV} adopts continuous, parameterized controls for translation and yaw.
Whereas prior work often adopts discrete action sets or retrieval-only evaluation, our benchmark emphasizes end-to-end control with an explicit stop action.
In terms of scale, \textsc{PairUAV} spans 72 scenes and 4.8M instances, approximately $22\times$ GeoText dataset, the largest prior aerial dataset in Table~\ref{tab:benchmark_compare}.
Finally, by centering on a drone agent and UAV-specific sensing geometry, \textsc{PairUAV} complements robot-centric ObjectNav datasets and enables evaluation under realistic aerial dynamics.

\begin{figure*}[!t]
    \centering
    \includegraphics[width=1.0\linewidth]{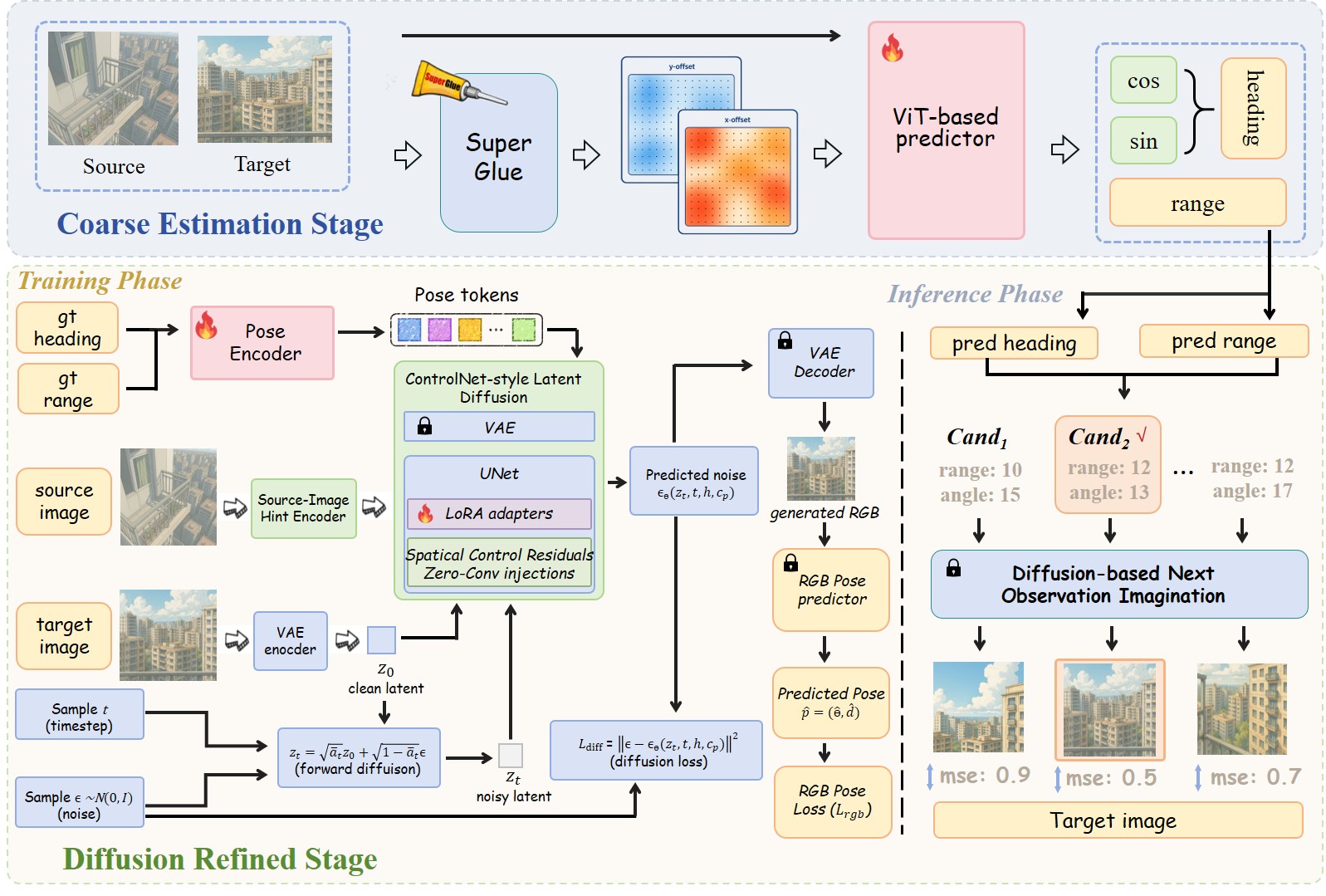}
    \vspace{-.3in}
    \caption{Overall architecture of our \textbf{DreamNav}. It consists of two successive stages, \ie, the first Coarse-Estimation Stage and the second Diffusion-Refined Stage. \textit{(Top)} In the Coarse-Estimation stage, given a source and a target image, we regress a coarse relative pose consisting of the heading angle and range from the source to the target. \textit{(Bottom)} In the Diffusion-Refined stage, we condition a ControlNet-style latent diffusion model on the source image through the ControlNet hint pathway and on the candidate pose through cross-attention tokens: during training, the ground-truth heading and range together with the source image are used to synthesize the target image via a diffusion loss; during inference, we feed the source image and a set of candidate poses, generate their corresponding views, and select the candidate pose whose synthesized view best matches the target image.}
    \label{fig:framework}
\end{figure*}

\section{Method}
\label{headings}

\subsection{Overview}
In this section, we study solutions to the last‑meter navigation problem for unmanned aerial vehicles. The task demands continuously parameterized control under strict accuracy constraints, making direct pose regression difficult under large motions and perspective changes and prone to poor generalization in novel scenes. To solve this problem, we introduce DreamNav, a feedback‑driven image‑based controller that targets metric‑accurate relative‑pose alignment under substantial illumination and viewpoint shifts. The overall architecture of DreamNav is shown in Figure~\ref{fig:framework}. Our approach consists of two stages: a coarse-estimation stage and a diffusion-refined stage. In Stage I, we represent yaw with coupled sin–cos outputs to sidestep periodic wrap‑around, and fine‑tune a Vision‑Transformer encoder to propose a coarse heading and range. In Stage II, we generate a set of local candidates centered around the initial coarse prediction. A diffusion-based next-observation generator synthesizes the observation for each candidate, and the candidate pose whose synthesized view best aligns with the target image is selected as the final prediction.

\subsection{Coarse-Estimation Stage}\label{Coarse-estimation}

Given a source image $S$ and a target image $G$ captured from the starting and target positions respectively, the coarse estimation stage predicts the relative pose from $S$ to $G$. We represent this navigation command as $p = (\hat{d}, \Delta\hat{\theta})$, where $\hat{d}$ denotes the forward range and $\Delta\hat{\theta}$ denotes the yaw rotation angle in the agent-centric frame. The estimation involves three main steps:

\paragraph{Displacement-Based Geometric Cue Extraction.}
For image-targeted navigation, prior works commonly concatenate the source image $S$ with the target image $G$ directly and fuse them with convolutional layers.
However, due to approximate translation equivariance~\cite{liu2018intriguing} of standard convolutions, such concatenation tends to discard absolute position information and thus fails to capture reliable cross-view spatial relations.
Therefore, we replace the original target image with the pixel-wise coordinate displacement from $G$ to $S$.

Specifically, we first establish sparse correspondences between $S$ and $G$ using a pretrained keypoint matcher (SuperGlue)~\citep{sarlin2020superglue}, yielding a set of $N$ matches 
\(
\mathcal{M}=\{(\mathbf{u}_i^S,\mathbf{u}_i^G,w_i)\}_{i=1}^{N},
\)
where $\mathbf{u}_i^S,\mathbf{u}_i^G\!\in\!\mathbb{R}^2$ denote the pixel locations in $S$ and $G$, and $w_i\!\in\![0,1]$ is the associated confidence. 
To reduce the impact of noisy or incorrect matches, we retain matches with confidence above a threshold $\tau$ and diffuse their displacements $\Delta\mathbf{u}_i=\mathbf{u}_i^G-\mathbf{u}_i^S$ into local neighborhoods using a Gaussian kernel $\kappa_\sigma(r)=\exp(-\tfrac{r^2}{2\sigma^2})$ with bandwidth $\sigma$, resulting in a dense displacement field through:
\begin{equation}
\begin{split}
D(\mathbf{p})
\;=\; &
\frac{1}{\alpha(\mathbf{p})}
\sum_{i=1}^{N}
w_i\,\kappa_\sigma\!\big(\lVert \mathbf{p}-\mathbf{u}_i^S\rVert_2\big)\,
\Delta\mathbf{u}_i,
\\
\alpha(\mathbf{p})
\;=\; &
\sum_{i=1}^{N}
w_i\,\kappa_\sigma\!\big(\lVert \mathbf{p}-\mathbf{u}_i^S\rVert_2\big)
\;+\;\varepsilon.
\end{split}
\label{eq:splat}
\end{equation}
where $\alpha(\mathbf{p})$ is a local normalization term, $\varepsilon$ ensures numerical stability, and $\mathbf{p}\in\Omega$ indexes image pixels.
In this way, $D(\mathbf{p})$ shrinks towards zero in poorly matched regions so that appearance cues prevail, whereas the displacement cue dominates in well-matched regions.

\paragraph{Dual-Cue Representation Fusion.}
The dense displacement field $D$ is then combined with the source image $S$ to form a dual-cue input, where the source image contributes appearance cues, while the displacement field contributes geometric cues.
We first normalize the displacement by image size $H\times W$ with an optional scalar $\lambda$:
\begin{equation}
\widehat{D}
\;=\;
\lambda\,
\begin{bmatrix}
D_x/W \\
D_y/H
\end{bmatrix}.
\label{eq:scale-norm}
\end{equation}
Meanwhile, the source image $S$ is normalized to $[-1,1]$.
Finally, the normalized source image and the normalized displacement map $\widehat{D}$ are concatenated along the channel dimension to form the final dual-cue representation $X\in \mathbb{R}^{H\times W\times 5}$, which serves as the input to the Vision Transformer.

\paragraph{Pose Regression Head.}\label{sec:head}
We attach a linear prediction head after the Vision Transformer to predict the initial relative pose $p = (\hat{d}, \Delta\hat{\theta})$, consisting of a forward range $\hat{d}$ and a yaw rotation angle $\Delta\hat{\theta}$.
To stabilize translation learning, we regress the range in a logarithmic parameterization that compresses its dynamic range while preserving fine resolution at short range.
The range loss is thus formulated as:
\begin{equation}
\mathcal{L}_{\text{dist}} = \Bigl(\hat{d} - \operatorname{sign}(d),\log\bigl(1 + |d|/\gamma\bigr)\Bigr)^2,
\label{eq:dist-loss}
\end{equation}
where $d$ is the ground-truth range and $\gamma$ is a scale factor.
Unlike range, however, rotation regression is complicated by the periodicity of angles, where geometrically identical rotations appear far apart under standard numerical regression losses.
To resolve this, we instead regress the sine and cosine of the rotation angle, yielding a smooth representation that avoids wrap-around discontinuities and stabilizes optimization.
The rotation loss is thus formulated as:
\begin{equation}
\mathcal{L}_{\text{rot}} = (\hat{s} - \sin \Delta\theta)^2 + (\hat{c} - \cos \Delta\theta)^2,
\label{eq:rot-loss}
\end{equation}
where $\Delta\theta$ is the ground-truth angle, while $\hat{s}$ and $\hat{c}$ denote the predicted sine and cosine components.
At inference, the rotation angle $\Delta\hat{\theta}$ is reconstructed from these components.
The total loss for the coarse estimation stage is a weighted sum of the range and rotation objectives:
\begin{equation}
\mathcal{L}_{\text{total}} = \mathcal{L}_{\text{dist}} + \lambda \mathcal{L}_{\text{rot}},
\label{eq:total-loss}
\end{equation}
where $\lambda$ is a balancing hyperparameter.

\subsection{Diffusion-Refined Stage}
\label{sec:refine}

While the coarse estimation stage provides a strong initialization, direct regression may still suffer from multi-modal visual ambiguity and local misalignment. To address this issue, we introduce a diffusion-based look-ahead verification stage. Instead of directly trusting the coarse pose prediction, we locally perturb the predicted heading-range pair, synthesize the visual consequence of each candidate pose, and select the candidate whose generated view best matches the target observation.

\paragraph{Heuristic Candidate Generation.}
\label{heuristic}

Given the coarse prediction $p_{\text{coarse}} = (\hat{d}, \Delta\hat{\theta})$ from the first stage, we construct a local search space around it. Specifically, we generate a $3 \times 3$ grid of candidate poses by applying fixed perturbations to heading and range. The offset sets are defined as
\[
\Omega_\theta = \{-10^\circ, 0^\circ, +10^\circ\}, \qquad
\Omega_d = \{-1.5, 0, +1.5\}.
\]
The candidate pose set is:
\begin{equation}
    \mathcal{P} =
    \Big\{
    \big(
    \hat{d} + \delta_d,\,
    \Delta\hat{\theta} + \delta_\theta
    \big)
    \;\Big|\;
    \delta_d \in \Omega_d,\,
    \delta_\theta \in \Omega_\theta
    \Big\}.
\end{equation}
This local grid allows the second stage to correct small errors in the initial pose estimate while keeping the search space computationally tractable.

\paragraph{Pose Information Encoding.}
\label{PoseRep}

To condition the diffusion model on each candidate pose, we encode the heading and range into compact pose tokens using a trainable pose encoder. The encoder is designed to preserve the periodic nature of heading angles and to provide a richer representation than raw scalar inputs.

For heading, we use sinusoidal features to avoid discontinuities at the angular wrap-around boundary:
\begin{equation}
\phi_\theta(\hat{\theta}) =
\big(
\cos(\omega_k \hat{\theta}),
\sin(\omega_k \hat{\theta})
\big)_{k=1}^{K_\theta}.
\end{equation}
For range, we first normalize the scalar value and then apply sinusoidal expansion:
\begin{equation}
\phi_d(\hat{d}) =
[\hat{d}]
\circ
\Big(
\cos(\nu_k \hat{d}),
\sin(\nu_k \hat{d})
\Big)_{k=1}^{K_d}.
\end{equation}
The final pose representation is obtained by concatenation:
\begin{equation}
\operatorname{PoseRep}(p)
=
[
\phi_\theta(\hat{\theta});
\phi_d(\hat{d})
].
\end{equation}
A trainable MLP maps this representation into pose tokens, which serve as cross-attention conditions for the diffusion model.

\paragraph{Next Observation Generation.}

We instantiate the next-observation generator as a ControlNet-style latent diffusion model. The source image $S$ is injected through the ControlNet hint pathway, which implements the spatial control condition. Specifically, $S$ is first processed by the input hint block $H_{\phi}(\cdot)$ to obtain a spatial hint feature:
\begin{equation}
    \mathbf{h}=H_{\phi}(S).
\end{equation}
Given the noisy latent $\mathbf{z}_t$, timestep $t$, pose token $\mathbf{c}_p$, and hint feature $\mathbf{h}$, the ControlNet branch produces multi-scale control residuals:
\begin{equation}
    \mathbf{r}_l =
    Z_l\big(C_l(\mathbf{z}_t, t, \mathbf{c}_p, \mathbf{h})\big),
\end{equation}
where $C_l(\cdot)$ denotes the $l$-th ControlNet block and $Z_l(\cdot)$ is a zero-convolution layer, implemented as a zero-initialized $1\times1$ convolution. The residual control feature $\mathbf{r}_l$ is added to the corresponding UNet feature at resolution level $l$, while an additional middle-block residual is produced at the bottleneck of ControlNet and added to the UNet middle block during denoising. In parallel, each candidate pose $p_k$ is encoded by the trainable pose encoder $E_p(\cdot)$ into cross-attention tokens:
\begin{equation}
    \mathbf{c}_p = E_p(p_k).
\end{equation}
The model denoises in the VAE latent space and decodes the resulting latent to obtain the candidate observation $\widehat{I}(p_k)$.

To preserve the pretrained generative prior while adapting the model to pose-conditioned UAV view synthesis, we freeze the VAE and most modules of the pretrained diffusion backbone. We fine-tune only a small set of task-specific parameters, including the trainable pose encoder, LoRA adapters in the ControlNet and UNet decoder/output blocks, and the source-image hint pathway of ControlNet. This design allows the source image to guide generation through multi-scale spatial control residuals, while the pose condition guides viewpoint-conditioned synthesis through cross-attention, without requiring full fine-tuning of the diffusion backbone.

\paragraph{Training.}

The diffusion model is trained with the standard denoising objective. Given a target image latent $\mathbf{z}_0$, a timestep $t$, and Gaussian noise $\epsilon$, the model receives the noisy latent $\mathbf{z}_t$ together with the source-image hint feature $\mathbf{h}$ and pose token $\mathbf{c}_p$. It learns to predict the added noise:
\begin{equation}
\mathcal{L}_{\text{diff}}
=
\mathbb{E}_{\mathbf{z}_0,t,\epsilon}
\left[
\left\|
\epsilon -
\epsilon_\theta(\mathbf{z}_t, t, \mathbf{h}, \mathbf{c}_p)
\right\|_2^2
\right].
\end{equation}

In addition, we introduce a frozen RGB pose predictor as an auxiliary supervision signal. 
Let \(F_{\text{rgb}}\) denote this frozen predictor. Given the generated denoised image \(\widehat{I}\), it predicts a heading-range pair:
\begin{equation}
(\hat{\theta}_{\text{rgb}}, \hat{d}_{\text{rgb}})
=
F_{\text{rgb}}(\widehat{I}).
\end{equation}
We then encourage the predicted pose from \(F_{\text{rgb}}\) to be consistent with the conditioning pose \(p=(\theta,d)\). The RGB pose-consistency loss is defined as:
\begin{equation}
\mathcal{L}_{\text{rgb}}
=
\lambda_\theta
\mathcal{L}_{\theta}
+
\lambda_d
\mathcal{L}_{d}
+
\lambda_n
\mathcal{L}_{n},
\end{equation}
where \(\mathcal{L}_{\theta}\) penalizes the heading discrepancy, \(\mathcal{L}_{d}\) penalizes the range discrepancy, and \(\mathcal{L}_{n}\) is a small normalization regularizer. The final training objective is:
\begin{equation}
\mathcal{L}
=
\mathcal{L}_{\text{diff}}
+
\lambda_{\text{rgb}}
\mathcal{L}_{\text{rgb}},
\end{equation}
where \(\lambda_{\text{rgb}}\) controls the overall weight of the RGB pose-consistency loss.

\paragraph{Inference.}

At inference time, we first obtain the coarse pose prediction from the first-stage pose predictor. We then construct the local candidate set $\mathcal{P}$ using the fixed heading and range offsets defined in Sec.~\ref{heuristic}. For each candidate pose, the diffusion model synthesizes a corresponding candidate observation. We compute the pixel-wise MSE between each synthesized view and the target image, and choose the candidate with the lowest MSE as the refined pose:
\begin{equation}
p^*
=
\operatorname*{argmin}_{p_k \in \mathcal{P}}
\left\|
\widehat{I}(p_k) - G
\right\|_2^2.
\end{equation}
The selected pose $p^*$ is used as the final refined prediction.

\section{Experiment}
\label{sec_exp}

In this section, we first introduce the dataset and evaluation metrics, and then present the experimental results and ablation studies. The implementation details of the DreamNav can be found in Appendix I.

\subsection{Experimental Setup}

\noindent \textbf{Dataset.} We evaluate our approach on the proposed PairUAV dataset. Following the partition protocol of University-1652 to ensure benchmarking consistency, we split the dataset into training and testing sets based on scenes. The training set encompasses 33 scenes with 701 buildings. To maximize the utilization of visual information, we perform exhaustive pairing for each building, using all $54 \times 54$ possible image combinations, which yields a substantial training corpus. The testing set consists of the remaining 39 unseen scenes containing 951 buildings. Similarly, we utilize all 54 images per building to generate navigation pairs, resulting in a large-scale evaluation benchmark of 2.0 million instances to rigorously assess the model's out-of-domain generalization.

\noindent \textbf{Metrics.} Following previous navigation works~\citep{anderson2018vision}, we report three standard metrics: success rate (SR), mean absolute angle error ($\mathrm{MAE}_H$), and mean absolute range error ($\mathrm{MAE}_R$). SR measures the fraction of evaluation episodes in which the vehicle, after executing its control sequence, comes to rest within $10m$ of the designated goal, which approximately corresponds to the accuracy limit of GPS. MAE is the average terminal position error across episodes, reflecting the typical deviation. The terminal position error is defined as the three-dimensional Euclidean distance between the UAV's final position and the target waypoint; higher SR and lower $\mathrm{MAE}_H$/$\mathrm{MAE}_R$ indicate better last-meter performance.


\begin{table}[!t]
    \centering
    \caption{Comparison with ObjectNav and Aerial Navigation methods. DreamNav achieves the best heading accuracy and the highest success rate (SR), demonstrating stronger navigation performance. Although our method does not obtain the lowest range error, heading estimation plays a more critical role in determining navigation success; therefore, the advantage in the more important $\mathrm{MAE}_H$ metric translates into a substantially higher final SR. $\mathrm{MAE}_R$ and $\mathrm{MAE}_H$ denote the Mean Absolute Error for Range and Heading, respectively. AVG is the average of the two.}
    \label{tab:main_result}
    \small
    \setlength{\tabcolsep}{3.5pt}
    \begin{tabular}{l|ccc|c} 
        \toprule                
        Methods & $\mathrm{MAE}_R \text{ (m)} \downarrow$ & $\mathrm{MAE}_H \text{ (deg)} \downarrow$ & AVG $\downarrow$ & SR (\%) $\uparrow$ \\
        \midrule                
        AI2THOR~\citep{7989381}    &  44.96    & 89.99 & 67.48 & 14.32 \\
        DINOv3-ViT7b~\citep{simeoni2025dinov3} & \textbf{15.77} & 89.86 & 52.81 &  10.33\\
        Sample4Geo~\citep{deuser2023sample4geo}    & 23.98    & 90.07  & 57.03 & 6.89\\
        \midrule
        Ours (Stage I)  & 29.52 & 40.29 & 34.91 & 19.81\\
        Ours (Stage II) & 29.16 & \textbf{38.78} & \textbf{33.97} & \textbf{23.51}\\
        \bottomrule             
    \end{tabular}
\end{table}


\subsection{Main Results}

\noindent \textbf{Baseline Methods.}
Given the absence of prior methods specifically tailored for image-goal aerial visual navigation with continuous parameter control, we benchmark against the representative approaches from three adjacent domains to establish a comprehensive evaluation:
\begin{itemize}
    \item \textbf{Vision-based servoing methods} with neural controllers and inputs comparable to ours, including AI2THOR~\cite{7989381}. 
    \item \textbf{Aerial visual localization methods} in similar UAV settings, including Sample4Geo~\cite{deuser2023sample4geo}, which is also built on the University-1652 benchmark~\cite{zheng2020university}.
    \item \textbf{Vision foundation models}, where we use DINOv3~\cite{simeoni2025dinov3} as a strong off-the-shelf visual representation baseline within the same controller.
\end{itemize}
All methods are benchmarked on the testing set of the PairUAV dataset. For the vision servoing models, we adapt their task setting to single-step action prediction and subsequently fine-tune their pre-trained backbones. For feature representation-based models, we first extract features using their officially released weights, and then add a classification head for training and evaluation.

\noindent \textbf{Quantitative Results.}
The comparison results on the PairUAV dataset are reported in Table~\ref{tab:main_result}.
Overall, DreamNav achieves the best average error and success rate among all compared methods, demonstrating its effectiveness for image-goal aerial navigation.
Although representation-based baselines such as DINOv3 and Sample4Geo obtain relatively low range errors, their heading errors remain close to $90^\circ$, leading to poor navigation success.
In contrast, our Stage-I model substantially reduces the heading error to $40.29^\circ$ and improves the success rate to $19.81\%$.
With the Stage-II refinement, DreamNav further improves the heading error to $38.78^\circ$, reduces the average error to $33.97$, and achieves the highest success rate of $23.51\%$.
Compared with the strongest baseline in terms of average error, DINOv3-ViT7b, DreamNav reduces AVG from $52.81$ to $33.97$, corresponding to a relative reduction of approximately $35.7\%$.
These results indicate that accurate heading estimation is crucial for reliable last-meter UAV navigation, and that our two-stage design provides more balanced and effective metric control than existing baselines.

\begin{figure}[!t]
    \centering
    \includegraphics[width=1.0\linewidth]{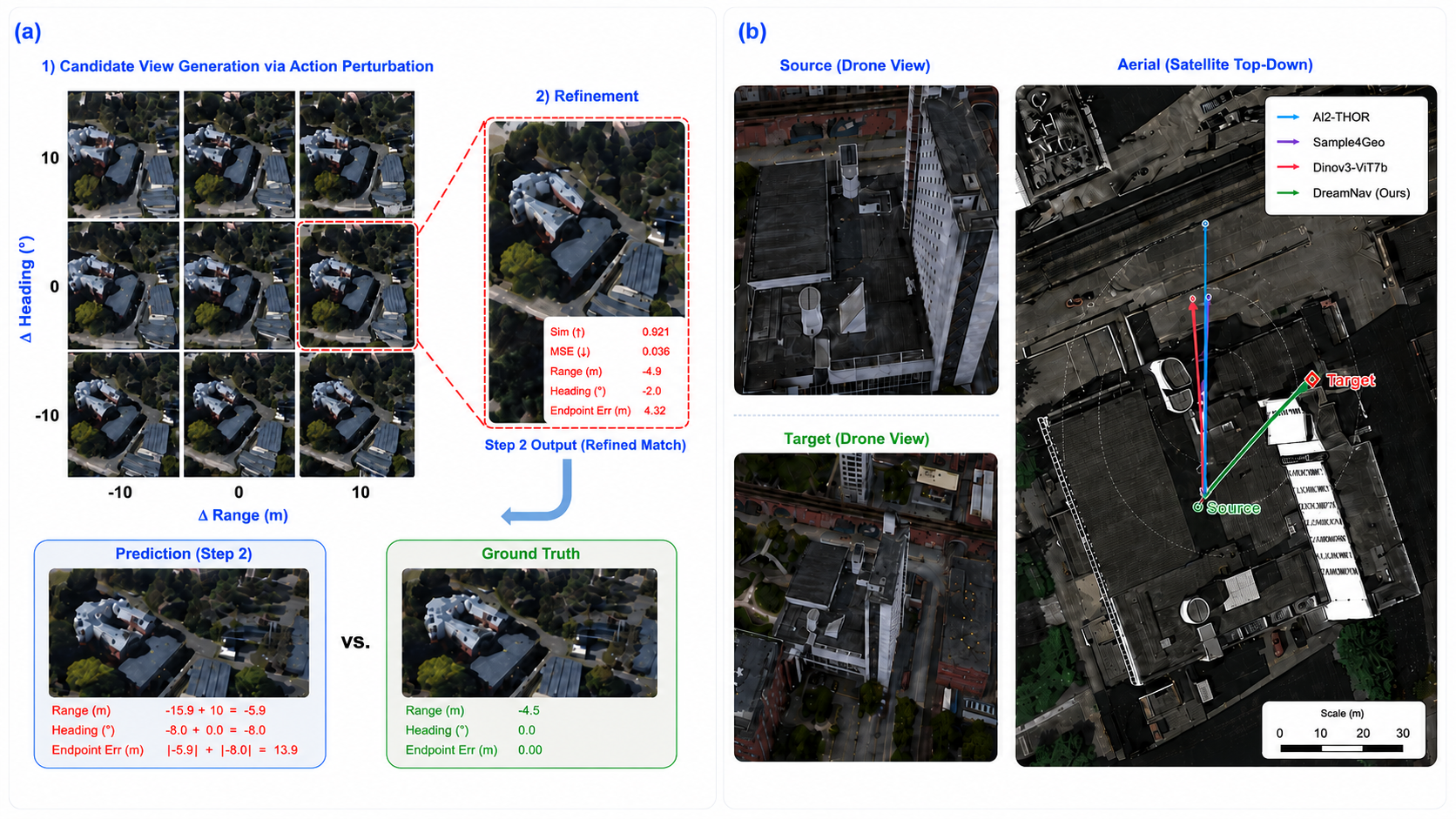}
    \caption{\textbf{Qualitative visualization of DreamNav.}
    (\emph{a}) Illustration of the two-stage inference process. In Stage I, we generate candidate next views by perturbing the predicted action in range and heading, where the $3{\times}3$ grid shows synthesized views under different conditioned offsets. The selected candidate is then refined in Stage II to produce a more accurate matched view, which is compared with the ground truth.
    (\emph{b}) Visual comparison with representative baselines. Given the source and target drone views, we project different methods' predictions onto the overhead satellite view. The red cross denotes the ground-truth target location, and colored arrows indicate the predicted navigation directions. DreamNav produces a trajectory closer to the ground truth, demonstrating more accurate last-meter localization.}
    \label{fig:vis_merged}
\end{figure}



\noindent \textbf{Qualitative Results.}
Beyond quantitative results, we provide qualitative visualizations on overhead satellite imagery to better analyze model behaviors.
For the Stage-II diffusion-based imagination module, we further conduct a qualitative analysis on the PairUAV dataset, as shown in Figure~\ref{tab:ablation},(a).
We randomly sample examples from PairUAV and used the coarse prediction from Stage-I as the central anchor to form a $3\times 3$ grid of conditioning inputs by applying systematic offsets to the heading and range.
It is shown that the synthesized next-views respond smoothly and consistently to these perturbations, where changing heading gradually rotates the rooftop orientation and facade layout, while moving forward/backward produces the expected scale change and parallax of nearby structures.
What's more, occlusion boundaries and relative depth ordering remain coherent across the grid, suggesting that the diffusion-refined stage captures geometry-aware viewpoint transformations rather than simply memorizing local textures.
This behavior verifies that Stage-II provides a controllable and physically plausible world model, which is crucial for reliable metric regression and navigation.

As illustrated in Figure~\ref{fig:vis_merged}\,(b), our method predicts a direction that is well aligned with the true goal, allowing the UAV to reach a location close to the target.
In contrast, existing baselines exhibit noticeable heading biases, leading to trajectories that drift toward incorrect orientations and accumulate large lateral errors.
This qualitative evidence further demonstrates that our trigonometric loss improves angular consistency, resulting in more accurate heading estimation and more reliable range regression.

\subsection{Ablation Studies} 

As shown in Table ~\ref{tab:ablation}, we observe that starting from a finetuned ResNet regression model as baseline (first row), introducing trigonometric regression yields the largest improvement, substantially reducing both range and heading errors (\eg, $\mathrm{MAE}_H$ drops from 84.36 to 68.30). 
We attribute this enhancement to our constructed smooth representation, which effectively mitigates angle-wrapping issues.
Adding dual-cue fusion further improves performance, with a pronounced gain in heading accuracy (68.30$\rightarrow$40.29 $\mathrm{MAE}_H$) and nearly doubling the success rate, suggesting that complementary geometric cues extracted from SuperGlue help disambiguate orientation under viewpoint changes. 
Even though incorporating the diffusion-refined stage brings marginal additional gains by itself, feeding first-stage outputs as conditions to the second-stage diffusion model (Action Enc.) further unlocks the benefit of second stage.
The full model (last row) achieves the best $\mathrm{MAE}_{AVG}$/$\mathrm{MAE}_H$ and the highest SR among all variants.

\begin{table}[!t]
  \centering
  \caption{Ablation studies on the test set of PairUAV. The baseline (first row) employs a finetuned ResNet. Trig. Reg., Dual-Cue, and Refine Stage. denote Trigonometric Regression, Dual-Cue Representation Fusion, and the Diffusion-Refined Stage, respectively.}
  \label{tab:ablation}
  \resizebox{\columnwidth}{!}{%
    \begin{tabular}{@{}ccc|cccc@{}}
      \toprule
      Trig. Reg. & Dual-Cue & Refine Stage &
      $\mathrm{MAE}_R \text{ (m)} \downarrow$ & $\mathrm{MAE}_H \text{ (deg)}  \downarrow$ & $\mathrm{MAE}_{AVG} \downarrow$ & SR (\%) $\uparrow$ \\
      \midrule
      & & & 31.34 & 84.36 & 57.85 & 6.00 \\
      \checkmark & & & 31.96 & 68.30 & 50.13 & 7.13 \\
      \checkmark & \checkmark & & 29.52 & 40.29 & 34.91 & 19.81 \\
      \checkmark & \checkmark & \checkmark & \textbf{29.16} & \textbf{38.78} & \textbf{33.97} & \textbf{23.51} \\
      \bottomrule
    \end{tabular}%
  }
\end{table}

\section{Conclusion}

In this paper, we introduce \textbf{DreamNav}, a coarse-to-fine framework for last-meter precision navigation of UAVs. We first propose a robust regression policy with trigonometric parameterization to mitigate optimization instabilities in rotation estimation. Additionally, we introduce a diffusion-refined stage that leverages a world model to hallucinate future observations, selecting the optimal trajectory through visual consistency. To facilitate rigorous evaluation, we present \textbf{PairUAV}, a large-scale benchmark covering diverse aerial scenes. Extensive experiments affirm the superior accuracy and zero-shot generalization of our method against strong baselines. We hope our work contributes to robust aerial visual servoing and inspires future research in autonomous drone agents.

\bibliography{main}
\bibliographystyle{iclr2026_conference}

\end{document}